\documentclass[letterpaper, 10 pt, conference]{ieeeconf}
\IEEEoverridecommandlockouts{}
{}
 {}
\usepackage{hyperref}
\usepackage{url}
\usepackage{graphics} 
\usepackage{epsfig} 
\usepackage{mathptmx} 
\usepackage{times} 
\usepackage{booktabs}	 
\usepackage{amsfonts}	 
\usepackage{nicefrac}	 
\usepackage{microtype}	 
\usepackage{algorithm, multirow, xcolor}
\usepackage{algorithmicx}
\usepackage{algpseudocode}
\usepackage{mathtools}
\usepackage{graphicx}
\usepackage{cases}
\usepackage{array}
\usepackage{color}
\usepackage{float}
\usepackage{amssymb}
\usepackage{hhline}
\usepackage{cite}
\usepackage{makecell}
\usepackage{tablefootnote}
\usepackage{wrapfig}
\usepackage{amsmath}
\usepackage{amsthm, amssymb}
\theoremstyle{definition}

\usepackage{soul}
\usepackage{graphicx}
\usepackage{subfigure}
\usepackage{epstopdf}
\usepackage[skip=5pt]{caption}
\usepackage{dblfloatfix}

\title{\LARGE \bf
SYNLOCO: Synthesizing Central Pattern Generator and Reinforcement Learning for Quadruped Locomotion
}

\author{Xinyu Zhang$^{1*}$, Zhiyuan Xiao$^{1*}$, Qingrui Zhang$^1$, and Wei Pan$^2$
 \thanks{$^*$These authors contributed equally to this work. }
 \thanks{$^{1}$School of Aeronautics and Astronautics, Sun Yat-sen University, Shenzhen 518107, P.R. China. Correspondence to: Qingrui Zhang ({\tt \small zhangqr9@mail.sysu.edu.cn})}
 \thanks{$^{2}$Department of Computer Science, The University of Manchester, UK and Department of Cognitive Robotics, Delft University of Technology, Netherlands.}
 }

\begin{document}

\maketitle
\thispagestyle{empty}
\pagestyle{empty}

\begin{abstract}

The Central Pattern Generator (CPG) is adept at generating rhythmic gait patterns characterized by consistent timing and adequate foot clearance. Yet, its open-loop configuration often compromises the system's control performance in response to environmental variations. On the other hand, Reinforcement Learning (RL), celebrated for its model-free properties, has gained significant traction in robotics due to its inherent adaptability and robustness. However, initiating traditional RL approaches from the ground up presents computational challenges and a heightened risk of converging to suboptimal local minima. In this paper, we propose an innovative quadruped locomotion framework, SYNLOCO, by synthesizing CPG and RL that can ingeniously integrate the strengths of both methods, enabling the development of a locomotion controller that is both stable and natural. Furthermore, we introduce a set of performance-driven reward metrics that augment the learning of locomotion control. To optimize the learning trajectory of SYNLOCO, a two-phased training strategy is presented. Our empirical evaluation, conducted on a Unitree GO1 robot under varied conditions—including distinct velocities, terrains, and payload capacities—showcases SYNLOCO's ability to produce consistent and clear-footed gaits across diverse scenarios. The developed controller exhibits resilience against substantial parameter variations, underscoring its potential for robust real-world applications.

\end{abstract}

\section{Introduction}
Quadruped robots are expected to be more common in industrial and daily life due to their extraordinary mobility competence in rough terrains and recent progress in computing technology. In particular, quadruped robots offer the best advantage in terms of mobility and versatility in various harsh and dangerous scenarios, {e.g.}, power plant inspections\cite{yoshiike_development_2017}, underground exploration\cite{tranzatto_cerberus_2022}, and planet exploration\cite{arm_scientific_2023}, {etc.}. In these challenging tasks, quadruped robots must traverse uneven terrain with various payloads. Hence, robust locomotion capabilities are highly expected for quadruped robots in real-world applications.

\begin{figure}[!t]
 \centering
 \includegraphics[width = \linewidth]{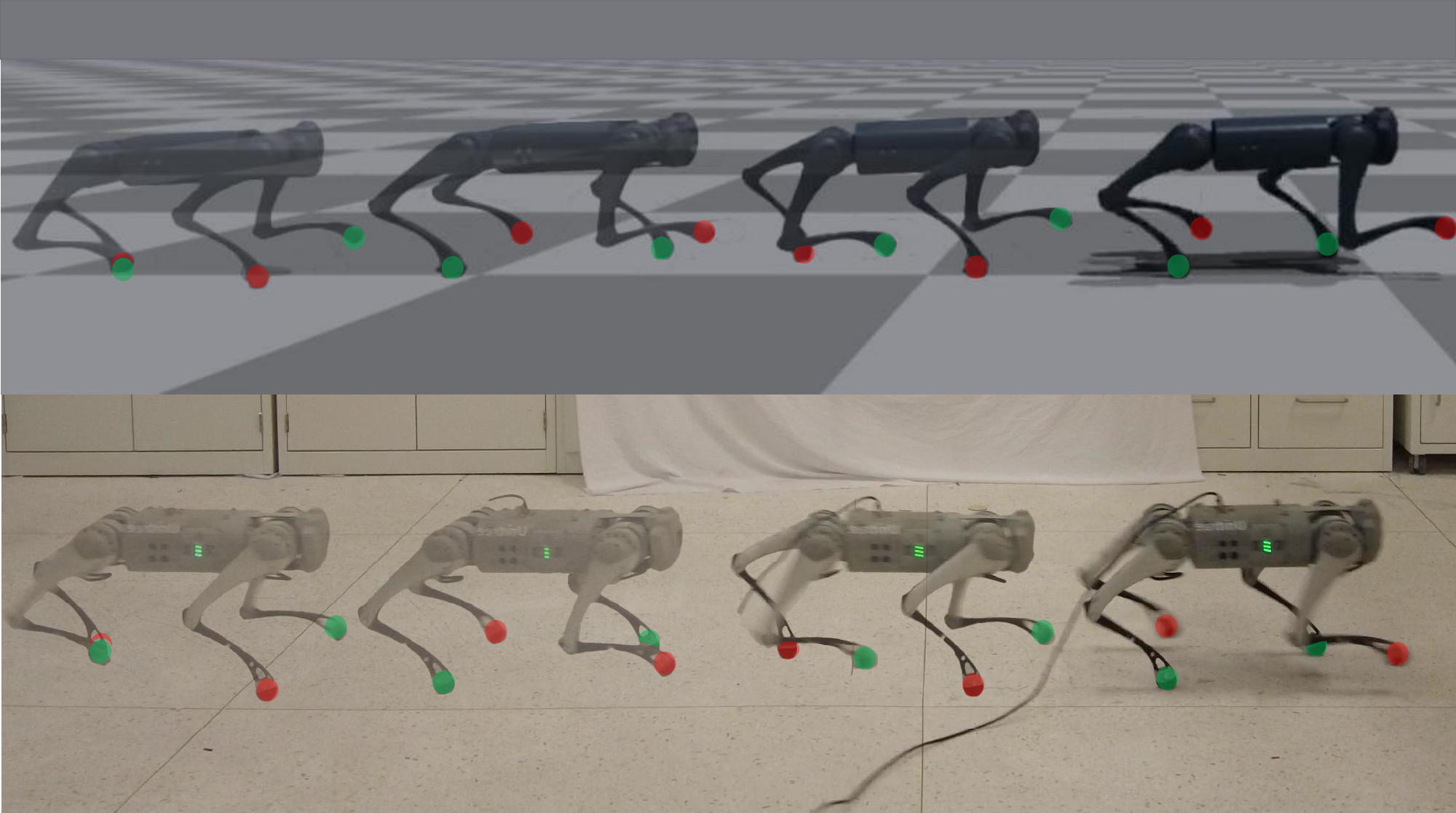}
 \caption{Training and evaluation snapshots. Top: Training in Isaac Gym. Bottom: Evaluation under a time-varying velocity command. }
 \label{fig: snapshot}
\end{figure}

An interesting solution to quadrupled locomotion is based on Central Pattern Generators (CPGs), which are circuits in animal self-contained integrative nervous systems to generate repetitive movements, for example, walking, swimming, crawling and flying \cite{yuste_cortex_2005,delcomyn_neural_1980}. Therefore, a CPG can generate rhythmic joint signals for locomotion control without using model information \cite{wang_cpg-based_2021,bellegarda_cpg-rl_2022,thor_fast_2019}. CPG-based locomotion control is simple, stable, and time-efficient, making it pleasant for real-time online implementations. However, most CPG-based locomotion control methods are open-loop, so they are barely adaptive to environmental changes. Parameter tuning is laborious and non-trivial for all CPGs, which often requires a great amount of expert knowledge or uses advanced optimization techniques, {e.g.}, genetic algorithms \cite{thor_fast_2019}. It should also be noted that CPG-based quadruped locomotion often performs poorly \cite{thor_generic_2021}.


Reinforcement learning (RL) does not need a mathematical model similar to CPGs, which recently has shown great potential in quadruped locomotion control \cite{hwangbo_learning_2019,lee_learning_2020,choi_learning_2023,miki_learning_2022}.
In RL, a locomotion control policy is learned using collected input and output data through robot interactions with environments \cite{hwangbo_learning_2019}. The learning process of RL is fully data-driven, so no model information is required. Policy learning in RL is a trial-and-error process that requires a large amount of data. Hence, a simulator is often necessary to generate enough data for training because real-world experiments are time-consuming, dangerous, and expensive. However, the policy learned in the simulation would experience performance degeneration due to the discrepancy between the simulated and real-world environments. Besides, mapping from sensory information to joint commands often results in nonintuitive motions. Designing valid reward functions for effectively learning natural locomotion under different poses and gaits is challenging. Specialized knowledge will always facilitate the reward design of RL.

\begin{figure}[tbp]
 \centering
 \includegraphics[width = \linewidth]{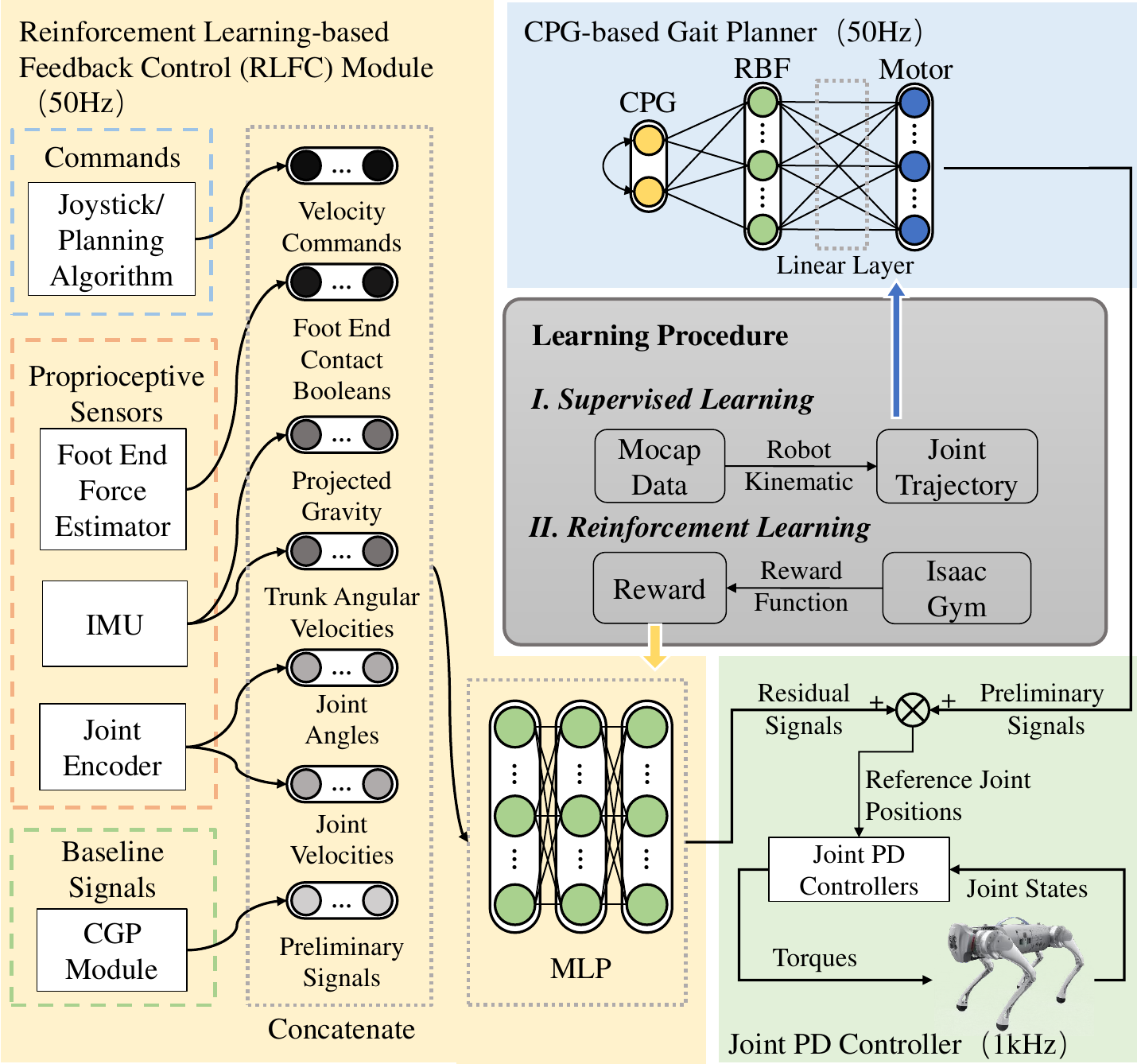}
\caption{ The SYNLOCO framework synthesizes RL and a CPG for quadruped locomotion.}
 \label{fig: framework}
\end{figure}

In this paper, we are interested in developing stable, natural, and robust locomotion control of quadruped robots in a model-free manner. A framework called SYNLOCO is proposed, which takes the merits of CPG and RL and synthesizes a CPG-based gait planner and a reinforcement learning-based feedback control (RLFC) module as depicted in Fig. \ref{fig: framework}. This framework enables a quadruped robot to react efficiently to various commands and environments with a steady and natural gait. Extensive experiments are performed to demonstrate the efficiency of the proposed framework. The whole contributions of this paper are three-fold:

\begin{itemize}
 \item[1)] 

 A quadruped locomotion control framework by synthesizing CPG and RL is presented, referred to as SYNLOCO. The proposed SYNLOCO framework can efficiently learn a stable, natural, and robust locomotion controller for a quadruped robot by leveraging the merits of both CPG and RL. The CPG in SYNLOCO generates stable rhythmic open-loop gait signals with sufficient foot clearance, which could be reference signals facilitating RL to avoid local minima. The RL module can learn a feedback control policy that dynamically modifies the unnatural gait signals of CPG using onboard proprioceptive sensors. Experiments illustrate that the proposed SYNLOCO can generate steady and robust locomotion with sufficient foot clearance in diverse situations, e.g., different velocity commands, terrains, and payloads.

 \item[2)] 

 The significantly enhanced velocity tracking performance and the robustness of the learned locomotion control policy are observed and found during RL controller by a plethora of performance-inspired reward functions, adding curriculum strategy and domain randomization. It is enlightening for the training setups of both our SYNLOCO algorithm and other RL-based quadruped locomotion control. Extensive experimental evaluations are performed to illustrate the advantages of learning a robust controller using the proposed approach. In particular, the quadruped robot can steadily march forward with $11$ kilograms of payloads that dramatically exceed the situation at the training stage, so it verifies the robustness of the learned locomotion control.



 \item[3)] 
A two-step training mechanism is proposed for SYNLOCO, which can facilitate fast, stable training via behavior cloning and reinforcement learning. In the first step, the CPG-based SYNLOCO planner is trained through supervised learning to clone the behavior from animal demonstration data. This step enables the CPG to learn from demonstration without interacting with the environment, accelerating the training process. Based on the trained baseline signal, the feedback RL policy is trained in a parallel way for generating residual signals for CPG. This two-step training mechanism would reduce the difficulty of policy learning in SYNLOCO, as the preliminary signal is. It is also favorable to learning natural locomotion as the CPG-based planner matches the behaviors of animals. 

\end{itemize}

The remainder of this paper is organized as follows. Section \ref{sec: Related Work} summarized the related works about CPG and RL. 
Section \ref{sec: Gait Planner} introduces the CPG-based gait planner and its behavior cloning method. Followed by the feedback controller and its RL training approach in section \ref{sec: Residual Reinforcement Learning}. The experimental results for our method are presented in Section \ref{sec: Results}. Conclusions are summarized in Section \ref{sec: Conclusion}.

\section{Related Works}\label{sec: Related Work}
\subsection{Model predictive control for quadrupled locomotion }
Quadrupled locomotion has been extensively investigated using model predictive control (MPC) \cite{mayne_model_2014,di_carlo_dynamic_2018,kim_highly_2019,bellicoso_dynamic_2018,grandia_perceptive_2023,kang_animal_2022}. In MPC, an online optimization problem is resolved recursively using predicted state trajectories on a finite time horizon by a robot model \cite{mayne_model_2014}. Therefore, an accurate model of the quadruped dynamics is indispensable for all MPC methods to generate stable or effective locomotion for quadruped robots. If the model parameters deviate far from the actual ones, the locomotion controller by MPC will perform poorly in practice. The heavy dependence of MPC on the model accuracy makes it difficult to apply to challenging uncertain situations, {e.g.} traversing uneven terrain \cite{lee_learning_2020}. MPC is also computationally expensive, so it is often challenging to meet real-time requirements \cite{fahmi_passive_2019}.

\subsection{Central  pattern generator for quadruped locomotion}
CPGs can produce rhythmic signals that are useful for the regulation of the repetitive movements of bio-inspired robots. Numerous artificial CPG models, such as the Hopf model \cite{gen_endo_experimental_2005}, the Kimura model \cite{ijspeert_swimming_2007}, and $SO(2)$ oscillator model \cite{thor_fast_2019}, have been proposed for locomotion control tasks. Among various CPG models, the $SO(2)$ oscillator model has received more attention because they are computationally efficient for online implementation \cite{thor_generic_2021,shi_reinforcement_2022, thor_versatile_2022}. However, parameter tuning remains difficult for all CPGs. 
Combining a CPG with other approaches has been discussed to build a general auto-tuning framework for locomotion. In \cite{thor_generic_2021}, a $SO(2)$ oscillator model is synthesized with a radial basis function (RBF) network that is optimized using a black-box method. A modular CPG-RBF network is introduced to realize legged robot control with fast learning \cite{thor_versatile_2022}. However, these works focus on the hexapod robot without feedback control and need better performance for quadruped robots. Bellegarda \emph{et al.} proposed a framework using an RL method to adjust CPG parameters \cite{bellegarda_cpg-rl_2022}. RL learns the gait pattern and foot end trajectory, which enables a quadruped robot to move with different foot clearance and body height. However, its foot end trajectory is predefined only for forward tasks. Shi \emph{et al.} proposed a framework that includes a learned CPG-based evolutionary trajectory generator with an RL-trained network to perform multitask locomotion\cite{shi_reinforcement_2022}. These auto-tuned CPGs cannot guarantee the learned gait pattern, since defining a reward function to specify the coupled relationship between legs is difficult. These works bring a viable approach combining CPG and other data-driven methods, inspiring us with a hybrid approach to design a controller. 

\subsection{Reinforcement learning-based locomotion control}
%
In recent years, RL has been successfully implemented to deal with quadruped locomotion \cite{rudin_learning_2022}. RL-trained networks demonstrated high-speed locomotion capabilities on deformable terrains\cite{choi_learning_2023}. Agile locomotion can be learned via DRL with an adaptive curriculum on commands and online system identification\cite{margolis_rapid_2022}. However, RL-based locomotion control requires a lot of data for policy training, which is too expensive for real experiments. The trial-and-error learning process of RL is also risky for real physical systems. Hence, policy training in terms of simulated data is preferred for robot control tasks \cite{rudin_learning_2022}. Considering the discrepancy between the simulated and real environments, this approach raises concern about the possibility of successful sim-to-real transfer. Peng \emph{et al.} realized the robot arm policy transfer by a domain randomization method that randomly changes the parameters of the model in training \cite{peng_sim--real_2018}. Tan \emph{et al.} analyzed the performance of domain randomization in quadruped locomotion \cite{tan_sim--real_2018}. Other methods include system identification \cite{hwangbo_learning_2019}. The domain randomization method allows the trained policy to be directly deployed to the robot without any other configuration, which is also applied in this paper. Despite the successful RL-based locomotion demos, learning an effective policy using an RL method is still challenging and expensive. The design of reward functions for natural and stable locomotion learning is still an open issue. In this paper, both issues will be addressed.

\section{CPG-based Gait Planner Module} \label{sec: Gait Planner}
The CPG-based gait planner is introduced to generate baseline gait signals for quadruped robots and optimized based on supervised learning as shown in Fig. \ref{fig: framework}. Similar to \cite{thor_generic_2021,thor_versatile_2022}, the CPG-based gait planner consists of a CPG layer, an RBF layer, and a motor layer. It is trained to produce desired gait patterns and frequencies via behavior cloning. The output of the gait planner is then used as the baseline gait signals to be regulated by the RLFC.

\subsection{CPG-based gait planner architecture}
 The CPG layer is a simple $SO(2)$ oscillator capable of generating stable rhythmic signals against perturbation. The $SO(2)$ oscillator has two neurons with a sigmoid activation function as in \eqref{eq:cpgWeight}.
\begin{equation}\label{eq:cpgWeight}
	\begin{gathered}
		\begin{bmatrix} 
 o_0(t+1) \\
 o_1(t+1) 
 \end{bmatrix}
		=\tanh\left(\alpha
		\begin{bmatrix} 
 \cos\phi & \sin\phi \\ 
 -\sin\phi & \cos\phi 
		\end{bmatrix}
		\begin{bmatrix} 
 o_0(t)	\\ 
 o_1(t) 
 \end{bmatrix}
		\right),
	\end{gathered}
\end{equation}
where $\phi$ is the desired oscillation velocity, $o_i(t)$ represents the state of neuron $i\in\left\{0,\;1\right\}$ at the  time instant $t$, $\alpha$ is a hyper-parameter set to be $0.01$. The parameter $\phi$ is set to $\frac{\pi}{60} $ to match the gait frequency of the demonstrated locomotion. 

The second layer is the Radial Basis Function (RBF) layer.
The RBF layer takes $o_0$ and $o_1$ as inputs and outputs a Gaussian-like expression.
\begin{equation}\label{eq:rbfActivation}
	R_h = exp{
	\left(
	-\frac{{(o_{0}-\mu_{h,0})}^{2}+{(o_{1}-\mu_{h,1})}^{2}}
	{\sigma^2_{RBF}}
	\right)},
\end{equation}
where $o_{i}$ are the CPG outputs, $\mu_{h,j}$ is the position, and $\sigma_{RBF}$ is the width of RBF neuron $h$. A bell-shaped curve $R_h$ is generated with a center $\mu_{h,n}$ and width $\sigma^2_{RBF}$. The center of the RBF neurons is uniformly distributed along the limit cycle of the CPG output. For node $h$, the center of dimension $n$ is denoted as $\mu_{h,n} = o_n \left( {hT}/{H} \right)$,
where $T$ is the signal period, and $H$ is the number of RBF neurons in the layer. In the RBF layer, the width of each function and the number of nodes jointly determined the output signals. With more RBF neurons and narrower widths, the signal can change rapidly, and \emph{vice versa}. We set $H=20$ and $\sigma_{RBF}=0.1$ for this layer. 

The motor layer is a fully connected neural network with one layer without activation functions. The output of the motor layer is a 12-dimensional vector representing the baseline position reference signals for 12 joints of the quadruped robots. The baseline signal indicates the desired gait pattern and frequency.

\subsection{Supervised Learning for behavior cloning}
The weights of the motor layer are trained via behavior cloning based on the supervised learning technique. Behavior cloning enables the CPG-based gait planner to clone gaits from demonstrations using supervised learning without interacting with environments. Animal demonstrations are obtained from \cite{peng_deepmimic_2018}, where a motion clip of an animal trotting at 1.5 Hz is selected as the demonstrated data. 
The foot end trajectories in the body frame are extracted from the raw data and adjusted according to the geometry information of the quadruped robot (Unitree Go1) used in this paper. 
The objective of the supervised learning is to minimize the mean square error (MSE) loss between the acquired foot end trajectories and generated trajectories from the CPG-based gait planner. The foot end trajectories of the CPG-based gait planner are calculated by robot forward kinematics. The training and validation set are split randomly from the acquired foot end trajectories by 7:3.

\section{Reinforcement Learning-based Feedback Control}
\label{sec: Residual Reinforcement Learning}
Following the idea of residual reinforcement learning, we use RL to learn a feedback control policy that generates residual control signals to modify the outputs of the CPG-based gait planner, thus resulting in the reinforcement learning-based feedback control (RLFC).  Details of the RLFC are provided later. 


\begin{table}[!t]
\centering
 \caption{Reward functions and weights}
 \label{table: Reward}
\begin{tabular}{lcc}
\toprule
Name&Expression&Weight\\
\hline
Linear velocity tracking&$\exp(-\frac{||v_{b,xy}^* - v_{b,xy}||^2}{0.25})$ &1dt\\
Angular velocity tracking&$\exp(-\frac{||\omega_{b,z}^* - \omega_{b,z}||^2}{0.25})$&0.5dt\\

Linear velocity penalty&$v_{b,z}^2$ &-2dt\\
Angular velocity penalty&$||\omega_{b,xy}||^2$&-0.05dt\\
Trunk orientation&$||\hat{g}_{x}||^2+||\hat{g}_{y}||^2$&-5dt\\
Trunk height&$1-\exp(-\frac{||h_b-h_b^*||^2}{8.1\times10^{-4}})$&-1dt\\
Joints acceleration&$||\frac{\dot{q}_{j-1}-\dot{q}_{j}}{dt}||^2$&$-1\times10^{-7}$dt\\
Action rate&$||q^*_{j-1}-q^*_{j}||^2$&-0.005dt\\
Self collision&$n_{collisions}$&-0.001dt\\
Foot air time&$\sum^4_{f=0}(t_{air,f}-0.5)$&1.5dt\\
Foot end position&$\sum^4_{n=1}\exp(-\frac{||p_{f}-p_{d}||^2}{0.02})$&0.3dt\\
\bottomrule
\end{tabular}
\end{table}




\textbf{\emph{Observations:}}
The observation space includes high-level commands $\{v^*_x,v^*_y,\omega^*_z\}$, trunk angular velocity $\omega_b$, unit gravity vector in body frame $\hat{g}$, joint positions $\{q_0, q_1,...,q_{11}\}$, joint velocities $\{\dot{q}_0,\dot{q}_1,...,\dot{q}_{11}\}$, the foot end contact booleans $\{c_0, c_1, c_2, c_3\}$, the actions at the last timestep, and gait planner signal. Some of the observations are resized before being sent to the network. The linear velocity command is scaled by 2.0. The angular rate command and trunk angular rates are scaled by 0.25. Joint velocities are scaled by 0.05.

\textbf{\emph{Actions:}}
The action space is 12-dimensional, the residual joint command signals for 12 joint actuators of a quadruped robot. The resultant position command signal $q_t$ for each joint is the sum of the CPG-based gait planner output $q_{CPG,t}$ and the RLFC module output $q_{RLFC,t}$, so $q_t = q_{CPG,t} +q_{RLFC,t}$.
A first-order low-pass filter filters the synthesized signal $q_t$ before sending it to the subsequent PD joint position controllers in the expression $s_t = \alpha q_t + (1-\alpha)s_{t-1}$, 
where $q_t$ is the origin signal and $s_t$ is the filtered signal in time step $t$ and $\alpha=0.7$. Joint PD controllers calculate joint torque sent to the motor and actuate the robot's locomotion. The subsequent proportional and derivative gains are $K_p=75$, $K_d=1.5$.
\begin{table}[bp]
\centering
\caption{PPO hyper-parameters}
 \label{table: PPO_paramters}
\begin{tabular}{cc}
\toprule
Parameter&Value\\
\hline
Learning rate&Adaptive\\
Batch size&98304 (4096$\times$24)\\
Mini-batch size&24576 (4096$\times$6)\\
Discount factor&0.99\\
GAE lambda&0.95\\
Desired KL-divergence&0.01\\
Entropy coefficient&0.01\\
Clip range&0.2\\
Number of epochs&5\\
\bottomrule
\end{tabular}
\end{table}

\textbf{\emph{Reward functions:}}
The total reward is a weighted sum of 11 terms shown in TABLE \ref{table: Reward}. The linear and angular velocity tracking rewards encourage a robot to track a target velocity. The trunk height and orientation rewards penalize the unsteady behaviors of the robot trunk. The rewards of joint acceleration and action rate penalize the dramatic change in the actual acceleration of the joint and the actions given. Self-collision term penalizes the collision between each foot and accelerates the training speed. The feet airtime term rewards the foot off the ground to avoid the foot end dragging the ground during its swing. The foot end position term penalizes the deviation of the foot end position during robot locomotion.

\textbf{\emph{RL policy:}} 
The classical RL algorithm, Proximal Probability Optimization (PPO) \cite{schulman_proximal_2017}, is selected to learn the feedback body controller in our method. A three-layer Multi-Layer Perceptron (MLP) with ELU activation functions will parameterize the actor and critic. The dimensions of the hidden layers are 512, 256, and 128, respectively. The remaining hyperparameters of PPO are listed in TABLE \ref{table: PPO_paramters}.

\textbf{\emph{Domain randomization:}}
Domain randomization techniques are employed in the training process to alleviate the sim-to-real gap issue. First, the dynamic parameters are randomized, including body mass and terrain friction coefficient, in each training episode to simulate the various environments. In addition, an instant linear velocity change toward the trunk is randomly applied to train a robust policy to withstand perturbation. A curriculum learning strategy similar to \cite{hwangbo_learning_2019} is introduced to steadily improve the locomotion robustness via gradually increasing the perturbation during the training. Suppose that the mean tracking reward reaches a predefined threshold, the impulse interval decreases, and the maximum impulse magnitude increases. Therefore, a fierce impulse is more likely to be applied to the robots. Finally, we add noise to the sensor's feedback to increase the controller's robustness against measuring errors and sensor faults. The parameters used for domain randomization are shown in TABLE \ref{table: Domain Randomization}. All randomized parameters follow a uniform distribution.
\begin{table}[tbp]
\centering
\caption{Parameters for domain randomization}
\label{table: Domain Randomization}
\begin{tabular}{clc}
\toprule
&Parameter&Range\\
\hline
\multirow{2}{*}{\makecell{Randomized\\dynamics}}
&Trunk mass&[-1,1] kg\\
&Foot end friction coefficient&[0.5,1.25]\\

\multirow{2}{*}{\makecell{Trunk\\impulse}}
&Trunk impulse magnitude $v_{xy}$&[-1.8,1.8] m/s\\
&Trunk impulse interval & 15 s\\

\multirow{4}{*}{\makecell{Sensor\\noises}}
&Trunk angular velocity noise&[-0.05,0.05] rad/s\\
&Gravity vector noise&[-0.05,0.05]\\
&Joint positions noise&[-0.01,0.01] rad\\
&Joint velocities noise&[-0.075,0.075] rad/s\\
\bottomrule
\end{tabular}
\end{table}

\textbf{\emph{Training setup:}}
Isaac Gym environment is used to conduct our training \cite{makoviychuk_isaac_2021}. As shown in Fig. \ref{fig: snapshot}, the quadruped robot is trained to follow high-level commands on both flat terrain and slope in parallel with 4096 agents. The simulation runs at 200 Hz, while the policy runs at 50 Hz. The maximum episode length is $20$ s (or $1000$ time steps). If the episode's duration reaches $20$ s  or the robot trunk collides with the ground, the episode is terminated and restarted. 
High-level commands include forward velocity $v^*_x$ ranging from [-1,1] m/s, lateral velocity $v^*_y$ ranging from [-1,1] m/s, and steering angular velocity $\omega^*_z$ ranging from [-1,1] rad/s. These commands are sampled from a uniform distribution every 10 seconds. At the beginning of each episode, the robot spawned in the air with the default pose and randomized high-level commands. The gait planner state is also initiated once the episode starts. 


\section{Experimental Results} \label{sec: Results}
\begin{figure*}[!t]
 \centering
 \includegraphics[width = \linewidth]{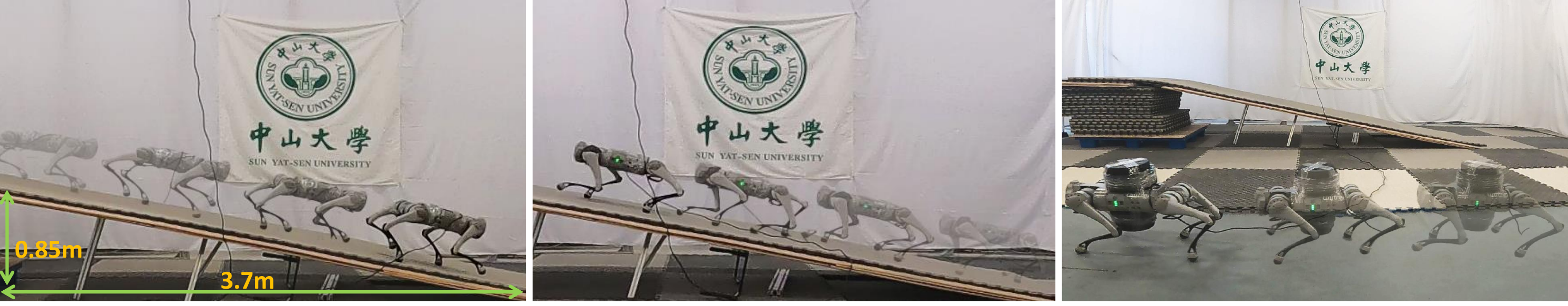}
  \caption{Experiments. Left: Downhill trotting. Middle: Uphill trotting. Right: Trotting with an 11 kg payload at 0.8 m/s.
}
 \label{fig: real experiment}
\end{figure*}

Real-world experiments are conducted to validate the efficiency of the proposed algorithm. The locomotion performance is evaluated by sending different forward velocity commands or in different terrains. Some snapshots of our experiments are shown in Fig. \ref{fig: real experiment}.

\subsection{Locomotion control for different velocity command}
\begin{figure}[tbp]
 \centering
 \includegraphics[width = \linewidth]{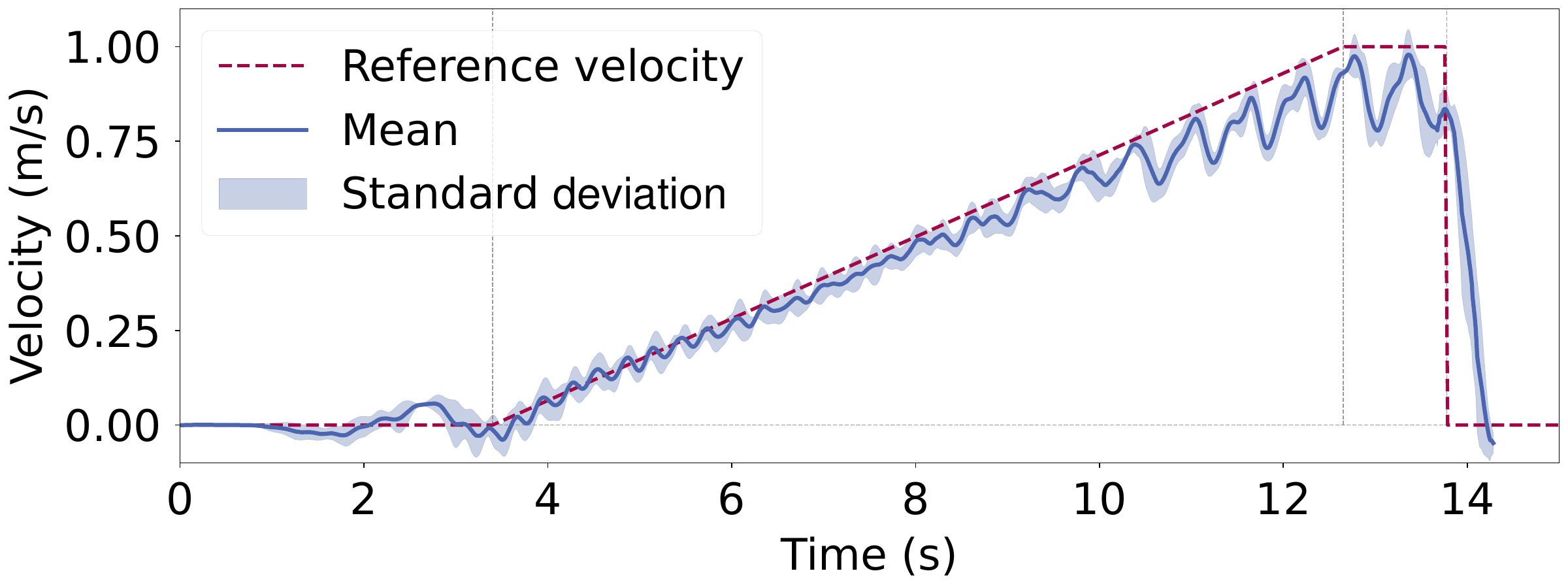}
 \caption{The velocity tracking performance. The experiment is repeated five times, based on which the mean and standard deviation are calculated.}
 \label{fig: Linear_vel_tracking}
\end{figure}

\begin{figure}[!ht]
 \centering
 \includegraphics[width = \linewidth]{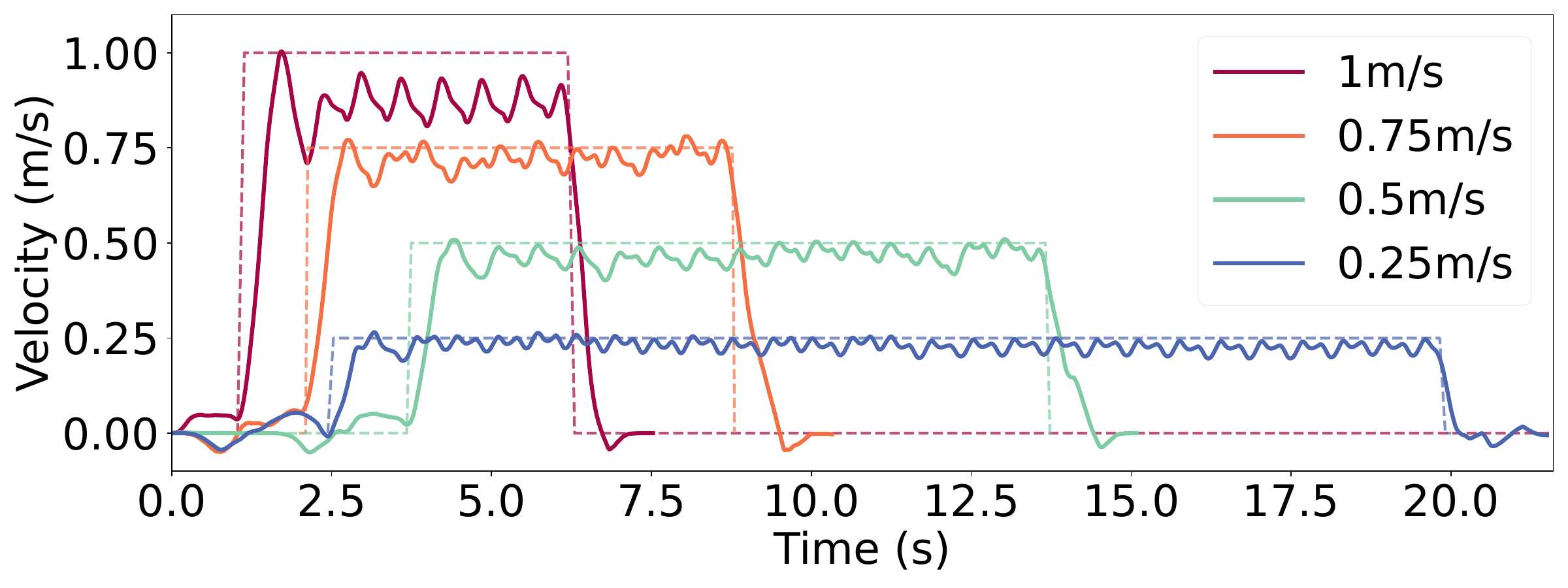}
 \caption{Trunk velocity responses under different commands.
}
 \label{fig: Fixed_vel_tracking}
\end{figure}

The velocity tracking performance under different commands was analyzed at first. 
To test its velocity-tracking performance, the robot is given a time-varying velocity command (from $0$ m/s to $1$ m/s). The velocity command $v^*$ will eventually remain at $1$ m / s for $1$ s. The OptiTrack motion capture system is used in the evaluation to obtain precise locomotion data for analysis. Fig. \ref{fig: Linear_vel_tracking} shows the velocity tracking performance. 
The trunk velocity tracked the command well at a slow velocity, but there were more oscillations at a higher velocity. This is because the robot trunk velocities are unavailable for our SYNLOCO algorithm. With the increase of velocity commands, the impact of the lack of velocity measurements will be more obvious. Hence, it implies that velocity measurements would be necessary for high-speed tracking. This conclusion is also verified via the testing at different constant velocity commands for the system, namely $0.25$ m/s, $0.5$ m/s, $0.75$ m/s, and $1$ m/s, respectively, as shown in Fig. \ref{fig: Fixed_vel_tracking}. The trunk height, pitching angle, and rolling angle responses demonstrate quadruped robots' steady, natural locomotion under our proposed design as shown in TABLE \ref{table: Trunk States}. 

\begin{table}[b]
\centering
\caption{Trunk states at different velocity commands}
\label{table: Trunk States}
\begin{tabular}{ccccccc}
\toprule
 
 \multicolumn{2}{c}{\multirow{2}{*}{Trunk state}}& \multicolumn{5}{c}{Velocity command (m/s)}\\
 \cline{3-7}
&&Varying&0.25&0.5&0.75&1.00\\
\hline

\multirow{2}{*}{\makecell{Velocity\\(m/s)}}&Mean&/&0.233&0.472&0.704&0.874\\
&Std. Dev.&/&0.014&0.020&0.033&0.040\\[1.25ex]

\multirow{2}{*}{\makecell{Height\\(m)}}&Mean&0.317&0.318&0.319&0.317&0.316\\
&Std. Dev.&0.005&0.004&0.005&0.006&0.007\\[1.25ex]

\multirow{2}{*}{\makecell{Pitch\\(deg)}}&Mean&2.745&1.872&2.754&2.622&3.702\\
&Std. Dev.&2.480&0.666&2.846&7.082&4.101\\[1.25ex]
\multirow{2}{*}{\makecell{Roll\\(deg)}}&Mean&-0.050&0.183&-0.162&-0.113&0.059\\
&Std. Dev.&1.361&1.371&1.289&1.192&1.514\\
\bottomrule
\end{tabular}
\end{table}

\begin{figure}[!ht]
 \centering
 \includegraphics[width = \linewidth]{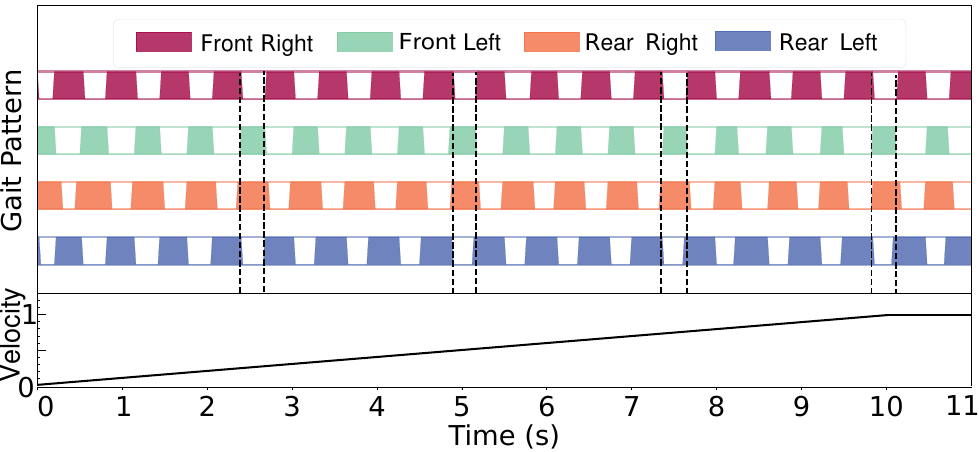}
 \caption{Robot gait pattern under a time-varying velocity command. The front right and rear left legs are in the same phase, which applies to the front left and rear right legs. It implies that the robot by SYNLOCO can keep a steady trotting gait with about 1.5Hz frequency under varying velocities.
}
 \label{fig: Giait Pattern}
\end{figure}

Fig. \ref{fig: Giait Pattern} shows the feet' contact patterns with the ground. The diagonal feet are in the same phase, while the adjacent feet are in the opposite phase. It implies that the robot can keep a steady trotting gait. The stance status (colored in the figure) is a more significant proportion than the swing status (blank in the figure), as shown in Fig. \ref{fig: Giait Pattern}. The experiment validated that the proposed SYNLOCO algorithm can generate continuous stable locomotion under different desired velocities with a desired fixed gait.

The actual foot end trajectories at a local frame are provided as shown in Fig. \ref{fig:  foot end local frame}. The robot can maintain the same foot clearance at different velocities. The foot clearances in the front legs are about 7cm, and the rear ones are about 4 cm. It is noted that the foot end is consistently fixed to the ground in its stance phase. The step length increases with increasing velocity. It implies that the locomotion under the proposed SYNLOCO algorithm is stable, steady, and natural, as illustrated in Fig. \ref{fig: snapshot}.


\begin{figure}[!ht]
 \centering
 \includegraphics[width = \linewidth]{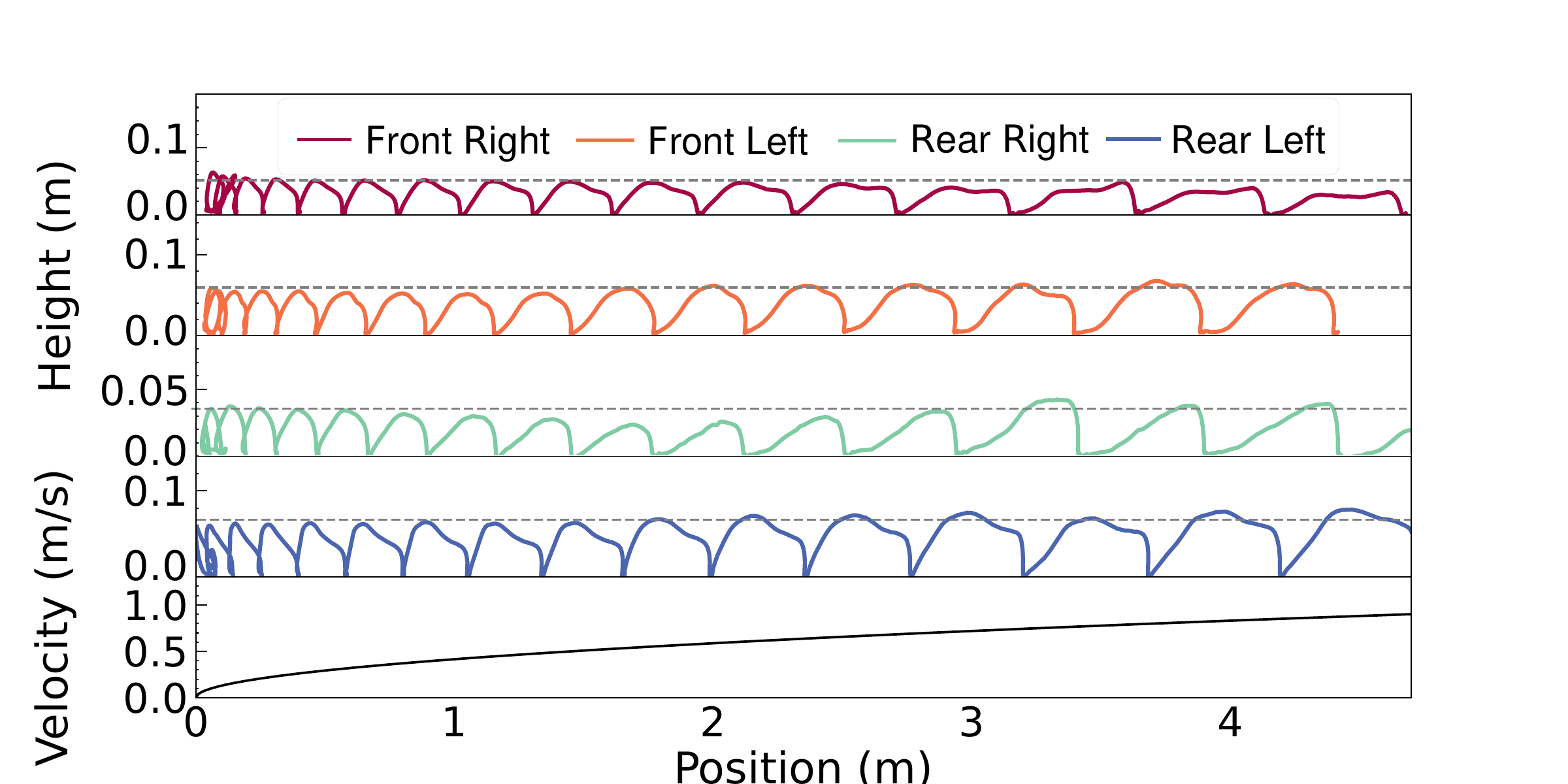}
 \caption{Foot trajectories in the XOZ plane of the local frame under a time-varying velocity. The dashed line represents the average foot clearance.
}
 \label{fig: foot end local frame}
\end{figure}

\begin{figure}[!ht]
\centering
\includegraphics[width = \linewidth]{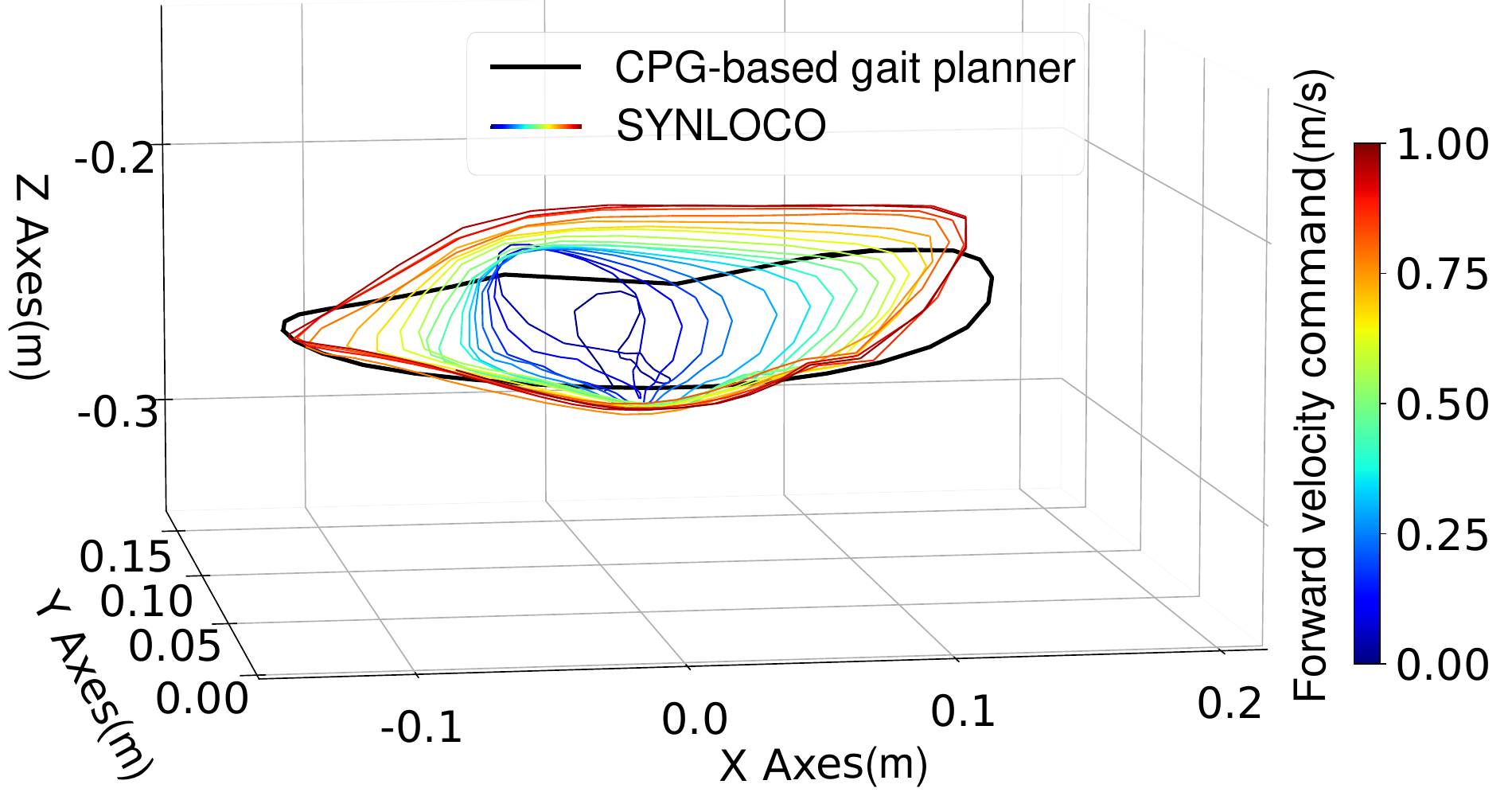}
\caption{Rear left foot trajectories under time-varying velocity command. Our SYNLOCO can adaptively regulate the gait signals while keeping sufficient foot clearance.}
\label{fig: foot end 3d trajectory}
\end{figure}

As shown in Fig. \ref{fig: foot end 3d trajectory}, the CPG-based gait planner generates a fixed foot baseline trajectory depicted in the black line. The baseline trajectory is regulated by the RLFC, leading to a varying trajectory as the colored line in Fig. \ref{fig: foot end 3d trajectory}. This implies that the RLFC can adjust the step length according to sensor feedback to keep foot clearance invariant. 

The relationship between the desired foot end trajectory and the actual foot end trajectory of the front right (FR) foot under different velocity commands is discussed as shown in Fig. \ref{fig:  foot end 2d trajectory}. Overall, the actual foot end trajectory resembles the desired ones with similar foot clearance and step length. Noticed that the desired trajectory is non-convex at the bottom, which is unmatched by the robot configuration, but the actual trajectory is smoothed. We can infer that the RL module learned how to generate feed-forward torque to compensate for the impulse from touching land by applying excessive modulation. The feedback controller controls the body pose by this method.

\begin{figure}[!ht]
 \centering
 \includegraphics[width = \linewidth]{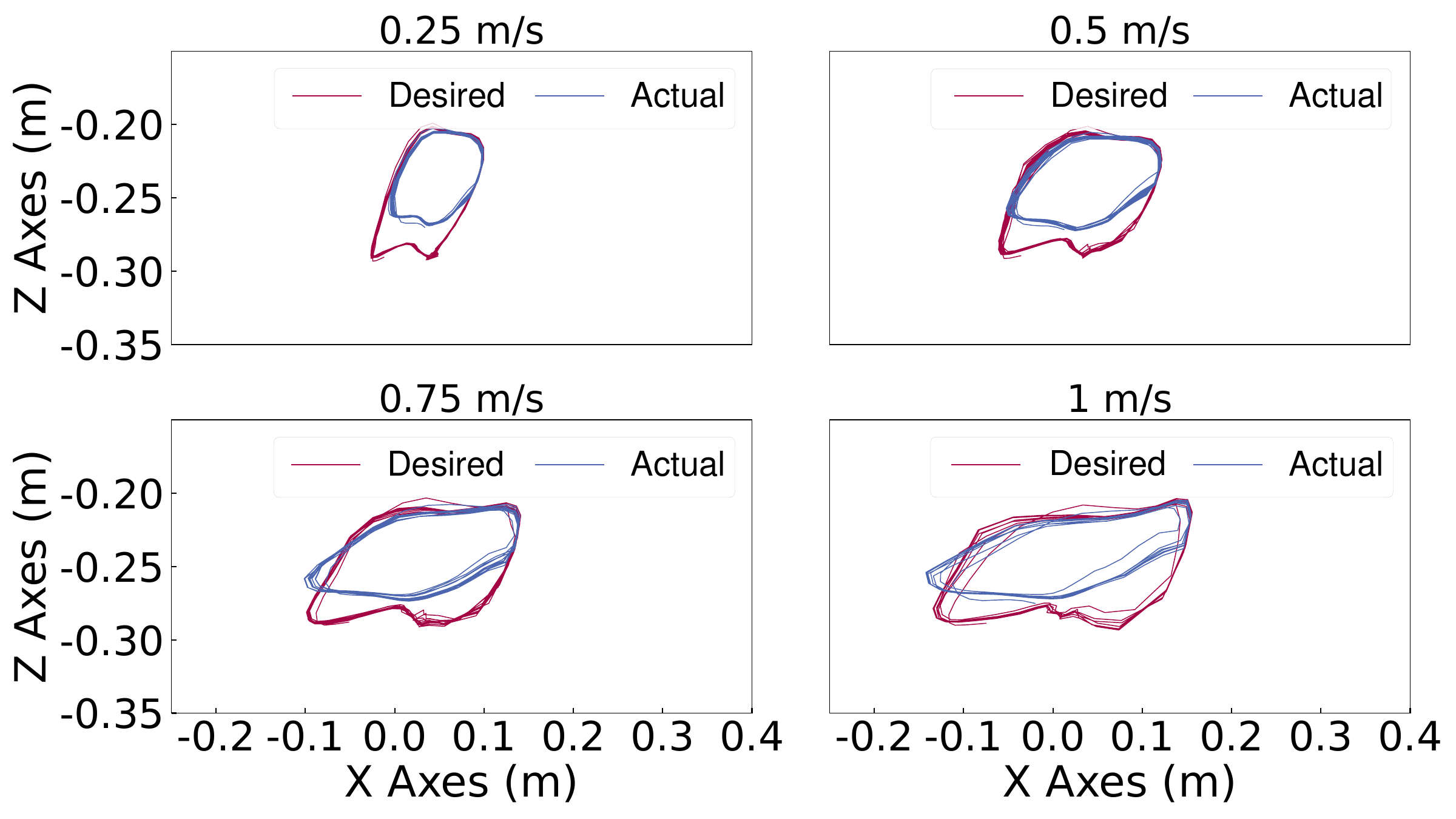}
 \caption{XZ trajectories of the front right foot in the body frame by our SYNLOCO algorithm at different velocity commands.
}
 \label{fig: foot end 2d trajectory}
\end{figure}

\subsection{Locomotion control for different tasks}
As shown in Fig. \ref{fig: real experiment}, the SYNLOCO algorithm is further evaluated in different environment configurations. Commands are sent using a joystick, including the forward velocity that ranges from $[0,1]$ m / s and the steering angular rate that ranges from $[-1,1]$ rad / s. The robot can perform stable and robust in different terrains with different payloads (Fig. \ref{fig: real experiment}). The transition between movements, such as marching forward and turning right, is steady and smooth. In addition, the robot can robustly traverse different surfaces, including rigid floors and deformable mattresses with various hardness.

It has also been demonstrated that the proposed SYNLOCO algorithm can allow the robot to steadily go uphill and downhill naturally, as shown in Fig. \ref{fig: real experiment}. The robustness of the proposed controller against parameter changes is also evaluated by making the robot carry an 11-kilogram payload. Such a payload is almost as heavy as the body mass of the Unitree Go1 robot. We dispersed dumbbell pieces above and below the robot trunk (placing 4.5 kg payloads at the bottom and 6.5 kg payloads at the top). Because the trunk mass varies in a small range ([-1,1] kg) by domain randomization during training, this experiment can illustrate our SYNLOCO controller has significant robustness against unexpected perturbations or varying model parameters.





\section{Conclusions} \label{sec: Conclusion}
This work presented a new quadruped robot controller, SYNLOCO, synthesizing a CPG-based gap planner module and a Reinforcement Learning-based Feedback Controller module. The simulation results showed that the controller can learn the quadruped robot locomotion under the desired gait pattern and frequency by our two-step training method. We also deployed the learned controllers in a real-world quadruped robot. The experiment results demonstrate that the learned controller can track different reference velocities and is robust for carrying heavy payloads and traversing uneven terrain. In SYNLOCO, the CPG-based gait planner can determine the baseline robot gait, and the RLFC module can adaptively regulate the baseline gait signals using sensor feedback.

The proposed SYNLOCO is only implemented to learn a stable and robust trotting gait. Therefore, one of the future works will be the extension of the proposed SYNLOCO algorithm to learn more gait patterns. Diverse terrains, such as stairs and gaps, will be considered in future work.

\bibliography{references}
\bibliographystyle{IEEEtran}

\end{document}